\RequirePackage{fix-cm}
\documentclass[smallextended]{svjour3}       

\usepackage[table]{xcolor}
 
\newcommand\MyBox[2]{
  \fbox{\lower0.75cm
    \vbox to 1.7cm{\vfil
      \hbox to 1.7cm{\hfil\parbox{1.4cm}{#1\\#2}\hfil}
      \vfil}%
  }%
}
\usepackage{float}
\usepackage{multirow}
\usepackage{booktabs}

\usepackage{subcaption}
\usepackage{cite}
\usepackage[tableposition=top]{caption}
\usepackage{diagbox}
\smartqed  
\usepackage{graphicx}

\usepackage{placeins}
\usepackage{changepage}
\journalname{AvoidingHelpAvoidance}
\begin{document}

\title{Avoiding Help Avoidance: Using Interface Design Changes to Promote Unsolicited Hint Usage in an Intelligent Tutor}

\titlerunning{Avoiding Help Avoidance}        

\author{Mehak Maniktala         \and
        Christa Cody \and
        Tiffany Barnes \and
        Min Chi
}


\institute{M. Maniktala          
           \and
           C. Cody
           \and
           T. Barnes\and
           M. Chi
           \at
              Department of Computer Science, North Carolina State University, Raleigh, North Carolina, USA  \\
              \email{mmanikt@ncsu.edu}  
}


\maketitle

\begin{abstract}
Within intelligent tutoring systems, considerable research has investigated hints, including how to generate data-driven hints, what hint content to present, and when to provide hints for optimal learning outcomes. However, less attention has been paid to \textit{how} hints are presented. In this paper, we propose a new hint delivery mechanism called ``Assertions" for providing unsolicited hints in a data-driven intelligent tutor. Assertions are \emph{partially-worked} example steps designed to appear within a student workspace, and in the same format as student-derived steps, to show students a possible subgoal leading to the solution. We hypothesized that Assertions can help address the well-known \textit{hint avoidance} problem. In systems that only provide hints upon request, hint avoidance results in students not receiving hints when they are needed. Our unsolicited Assertions do not seek to improve student help-seeking, but rather seek to ensure students receive the help they need. We contrast Assertions with Messages, text-based, unsolicited hints that appear after student inactivity. Our results show that Assertions significantly increase unsolicited hint usage compared to Messages. Further, they show a significant aptitude-treatment interaction between Assertions and prior proficiency, with Assertions leading students with low prior proficiency to generate shorter (more efficient) posttest solutions faster. We also present a clustering analysis that shows patterns of productive persistence among students with low prior knowledge when the tutor provides unsolicited help in the form of Assertions. Overall, this work provides encouraging evidence that hint presentation can significantly impact how students use them and using Assertions can be an effective way to address help avoidance.

\end{abstract}

\section{Introduction}
Studies suggest that hints, when provided appropriately, can augment students’ learning experience \cite{bunt2004scaffolding,puustinen1998help} and improve their performance \cite{bartholome2006matters}.  However, students may not use hints optimally \cite{duong2013prediction,aleven2006toward}; some abuse hints to expedite problem completion, and some avoid seeking help when they are in need \cite{aleven2000limitations, price2017factors}. Our goal is to redesign the hint interface to solve this help avoidance problem. 
Considerable research has investigated hints from several perspectives, including hint generation \cite{barnes2008pilot, price2016generating}, adaptive hint content \cite{ueno2017irt,conati2013understanding, kardan2015providing},  student help-seeking behavior \cite{aleven2006toward,price2017factors}, and hint timing \cite{razzaq2010hints}. However, few studies have specifically investigated how hint interfaces could reduce help avoidance (e.g. \cite{kardan2015providing, moreno1999cognitive}). 

Most intelligent tutoring systems (ITSs) provide solicited hints \emph{on-demand}, i.e, upon student request \cite{vanlehn2006behavior}. Other tutors try to circumvent help avoidance by providing \emph{unsolicited} hints when the system ``determines" they are needed, for example, after a long period of inactivity \cite{fossati2010generating}. However, students often ignore these unsolicited hints \cite{muir2012analysis,conati2013understanding}. In this work, we designed a new interface for unsolicited hints, called \textit{Assertions} to address this issue, and compared its impact on student learning outcomes with that of \textit{Messages}, text-based unsolicited hints that appear after student inactivity. The ultimate  goal of our research is to combine the new Assertions interface with a data-driven method to determine when providing an unsolicited hint would be most beneficial and least disruptive for students.

Our Assertions interface was designed based on user experience and multimedia design principles, including contiguity \cite{moreno1999cognitive}, attention \cite{healey2012attention}, expectation \cite{summerfield2009expectation}, and persuasion \cite{dillard2013affect, cialdini2009influence}. First and foremost, we hypothesized that placing Assertions \textit{contiguously} within the area of student \textit{attention} would make unsolicited hints more noticeable. Second, we believed students could more quickly interpret Assertions based on the \textit{expectation} set by formatting them like other problem-solving steps. Finally, we used \textit{persuasive} language asking students to use the Assertions as problem-solving subgoals. These features help Assertions act as partially-worked example steps, so they may garner the same benefits of worked examples, that have been shown to improve learning efficiency \cite{mclaren2008assistance}. We hypothesized that Assertions would reduce help avoidance for all students, by increasing the percentage of times help was received when it was needed. Further, we hypothesize that Assertions would have an aptitude-treatment interaction effect,  fostering productive persistence and improving posttest performance, among students with low prior proficiency. Persistence during training that leads to mastery of a subject or positive posttest outcomes is called productive persistence \cite{kai2018decision}.
 
The main contribution of this work is a principled design for a hint interface, Assertions, and a study to show that Assertions can be used to significantly reduce help avoidance for all students through interface alone. Our new proposed Assertions appear as partially-worked example steps, reducing the barriers to help usage while leveraging benefits of worked examples. The second contribution of this work is a new cluster-based method that combines posttest performance, effort (to quantify persistence), and unsolicited hint usage to discover productive persistence. Based on these clusters, we were able to show that students with low prior proficiency who received Assertions exhibit productive persistence. Since Assertions are automatically provided to students, they can be thought of from two perspectives: either as unsolicited hints, or as partially worked example steps. Therefore, in our related work and design sections below, we discuss Assertions from both of these perspectives.

\section{Related Work}
\subsection{Hints in ITSs}
ITSs have the unique ability to offer individualized help and feedback. While research suggests that such individual adaptation can significantly improve learning, they can take a considerable time to construct \cite{murray2003overview}. Example-based authoring tools such as CTAT often employ examples to construct production rules, where teachers work problems, predict frequent incorrect approaches, and then develop rules and manually write appropriate hints \cite{koedinger2004opening}. Such systems have also been augmented by methods such as Bootstrapping Novice Data, where they use data to build initial models, but still require manual expert hint authoring \cite{mclaren2004bootstrapping}. However, considerable time must still be spent on identifying student approaches for hint generation and to write the messages that appear with hints.

Data-driven assistance saves time and resources by reducing the need for an expert -- typically such systems include a hint template, authored once for all hints, and use algorithms to generate data to include in the template-based hint. Data-driven methods have increasingly been used to generate personalized help in intelligent tutors for open-ended multi-step problem-solving domains. Barnes and Stamper invented the  Hint Factory, the first data-driven method to generate next-step hints, and demonstrated its effectiveness for propositional logic \cite{stamper2008hint, barnes2011using}. The approach uses prior students' transaction log data to form an interaction network and then runs the Bellman backup for value iteration to score problem-solving states (snapshots of an on-going or completed problem solution attempt). To provide hints, the Hint Factory finds states matching students’ current work and delivers hints using the next reachable state with the highest score. Barnes and Stamper and their colleagues did pioneering work to explore program representations that could be used for hint generation for novices \cite{jin2011towards} and demonstrated how they could be used to provide hints for 80\% of the states in a historical dataset \cite{jin2012program}. Similar to the Hint Factory, Fossati et al. \cite{fossati2015data} devised the Procedural Knowledge Model (PKM), that uses students’ global problem-solving behaviors to generate data-driven feedback for the iList tutor for programming with linked lists. Price, Barnes, and colleagues extended the Hint Factory approach to generate data-driven hints for novice programming \cite{price2016generating, price2017isnap, price2018impact}. Later, Paaßen et al. created the continuous hint factory to allow for hint generation for previously unobserved states \cite{paassen2017continuous}, while Price et al. devised the SourceCheck algorithm that leveraged similar representations to generate hints based on a set of student solutions rather than the trace data that the original Hint Factory uses \cite{price2017evaluation}.  Rivers et al. developed a data-driven hint generator for ITAP (Intelligent Teaching Assistant for Programming) that uses a similar set of tools including state abstraction, path construction, and state reification to generate personalized hints \cite{rivers2017data}. This method extends the Hint Factory by enhancing the solution space and creating new edges for states that are disconnected. This allows the ITAP method to generate hints even for states that are not present in the prior data. For this work, we extended  the Hint Factory to provide personalized hints for logic with 100\% availability as described in Section \ref{sys}.

Aleven et al. have shown that students often display poor help-seeking behaviors within intelligent tutors, including \textit{help avoidance}, where students \emph{could} benefit from seeking help but choose not to, and \textit{help abuse}, where students use help excessively when they could solve a problem without assistance \cite{aleven2006toward}. Studies by Price et al., Almeda et al., and Roll et al. have confirmed that help avoidance is pervasive across domains and systems with students ignoring hints \cite{price2017hint,almeda2017help, roll2006help}. In one study, Roll et al. showed that meta-cognitive feedback improved student’s help-seeking skills but did not affect their domain learning \cite{roll2006help}. Price et al.’s research study on help-seeking by novice programmers showed that students have several reasons for not requesting on-demand hints, including uncertainty about whether system help would be useful, or a desire to be independent \cite{price2017hint}. Some tutoring systems prevent help avoidance by providing \emph{unsolicited hints} rather than relying on student help-seeking through ``on-demand” hint requests \cite{arroyo2001analyzing,murray2006comparison, marwan2019impact}. Arroyo et al. \cite{arroyo2001analyzing} and Murray et al. \cite{murray2006comparison} showed that unsolicited hints promoted learning gains for a subset of students. However, a study by Muir and Conati showed that students often ignore unsolicited hints \cite{muir2012analysis}.

Several studies have tried to encourage students to use unsolicited help by changing its content or placement. For example, Cody et al. showed that unsolicited, data-driven hints were more likely to be used if their content focused on next-step hints rather than more abstract, high-level hints \cite{cody2017investigating}. Conati et al. used eye-tracking to show that factors such as hint timing, and student’s attitude and prior knowledge can affect students’ attention towards unsolicited hints in a number factorization game \cite{conati2013understanding}. Kardan and Conati showed that unsolicited hints with tailored hint content along with highlighting and proximal hint placement improved student learning in a controlled study with AI SPACE \cite{kardan2015providing}. 

Despite their potential benefits, we argue that attempting to understand or use hints, and especially unsolicited ones, can increase students’ cognitive load while learning new concepts within a tutoring system. This is because students have to mentally integrate several sources of information, including on-demand hints, unsolicited hints, and the student’s own current solution attempt. Adding to this is the fact that, in many existing tutoring systems, the hints and the student solution workspace are physically located in different areas of the interface. As a result, we believe that by physically integrating those sources of information together, Assertions naturally reduce students’ working memory load and thus would facilitate student learning by accelerating the changes in their long term memory associated with schema acquisition \cite{sweller1988cognitive, sweller1999instructional}.

\subsection{Worked Examples}
Since we posit that Assertions can be seen not only as unsolicited hints, but from another perspective as partially-worked examples for single problem-solving steps, we discuss impacts of worked examples here. Extensive research has shown that worked examples, i.e. showing step-by-step problem solutions, can be as effective as problem solving to learn the same content yet the former generally need much less time \cite{mostafavi2015data, mclaren2008assistance}. In our prior work, we have added whole-problem worked examples to our tutor to help students learn the problem interface and problem-solving skills. In \cite{mostafavi2015data}, we found that the students who received data-driven worked examples were much more likely to complete the tutor, and did so in less time \cite{mostafavi2015data}. In another study \cite{shen2016reinforcement}, we found that when we use reinforcement learning (RL) to determine when to present whole-problem worked examples, the slow learners provided based on this RL policy had a significantly higher learning gains than their peers who received worked-examples at random. Further, our results from study on worked examples in Deep Thought \cite{liu2016combining} show that whole-problem worked examples benefit students early in the tutoring, but are comparable to hint-based scaffolding. We also observed that worked examples were less beneficial later in the tutoring sessions for lower proficiency students. Our work with Pyrenees, a probability tutor, suggests that step-level Worked Examples can also promote learning \cite{zhou2015impact}. This work suggests that students do not resist following these step-level worked examples, that are essentially unsolicited hints provided in student workspace.

One mechanism proposed by Sweller et al. for the success of worked examples is through reduction in the cognitive load when students are learning new concepts \cite{sweller2006worked}. Their work discusses the principles underlying cognitive load theory and how worked examples reduce the need for learners to engage in inference processes which might otherwise require heavy demands on students’ working memory. On the other hand, much prior work found that asking students to \textit{justify} their solution steps, referred to as self-explanations, can greatly improve their learning \cite{chi1994eliciting, conati2000toward, aleven2004evaluating}. Furthermore, asking students to explain expert-designed worked examples can be more effective than problem solving alone \cite{chi1988learning, weerasinghe2002enhancing}. For example, Weerasinghe and Mitrovic explored the impact of self-explanations in KERMIT-SE, a tutor for the open-ended domain of database design. They engaged students in tutorial dialogues upon errors in solutions and found that it improved student performance in both conceptual and procedural knowledge \cite{weerasinghe2002enhancing, weerasinghe2004supporting}. In this work, we design our new Assertions hint interface to act as expert-designed partially-worked example steps with self-explanations. However, there are two key differences between our work and that by Weerasinghe and Mitrovic: Assertions are provided to guide students on the next step instead of the current step, and they are provided after correct steps instead of incorrect steps. As described in section \ref{sys}, Assertions provide students with the content of a useful step, but students must provide an explanation before they can use the hint content in their solutions.

\subsection{Aptitude-Treatment Interaction}
Prior research in instructional strategies has shown the existence of aptitude-treatment interaction (ATI), where certain students are more sensitive to variations in the learning environment compared to less sensitive students who perform regardless of the treatment \cite{cronbach1977aptitudes,snow1991aptitude}. Researchers have explored the complex relationship between student aptitude and their interaction with unsolicited help. While Razzaq et al. found that students learned more reliably with hints they requested than unsolicited hints  \cite{razzaq2010hints}, Arroyo et al. observed higher learning gains for low performing students when unsolicited hints were provided \cite{arroyo2001analyzing}. Further, Murray et al. found that unsolicited help avoided the negative effects of frustration and saved students time when they were struggling  \cite{murray2006comparison}. Muir and Conati showed that students with low prior knowledge are likely to need hints the most, but they do not look at the hints as often \cite{muir2012analysis}. Kardan and Conati found that changes in unsolicited hint content and interface had a more pronounced effect on learning for students with lower initial knowledge \cite{kardan2015providing}. Similar to these studies, we hypothesize that an improved interface for unsolicited hints can increase hint usage and outcomes, especially for students with low prior knowledge. In this work, we believe that students whose initial tutor performance is lower may need more assistance to develop strategies for solving logic proofs, and therefore, may benefit more from an improvement in the hint interface. 

\subsection{Productive Persistence}
Recently, there is an increased interest in non-cognitive skills like persistence and self-control within education research \cite{kai2018decision}. Task persistence is defined as the continuation of a task despite difficulty. To quantify persistence, researchers used metrics of \emph{effort} \cite{dicerbo2014game}.  However, not all persistence is productive, Beck and Gong \cite{beck2013wheel} define unproductive persistence or ``wheel spinning" as when a student spends an excessively long time struggling to learn a topic without achieving mastery. They showed that if a student did not master a skill in ASSISTments (an online math learning platform) or the Cognitive Algebra Tutor in a reasonable amount of time, the student was likely to struggle and never master the skill. Their work presents connections between wheel-spinning and negative student behaviors such as disengagement and gaming, as well as recommendations to improve ITS design to address these issues. Research by Nelson et al. is well-known for their heuristic model of the help-seeking process where they suggest that unproductive persistence may be associated with help avoidance \cite{nelson1981help}. Studies suggest that the persistent effort that lead to mastery of a topic is \textit{productive persistence} \cite{kai2018decision}, and is often associated with short-term outcomes like improvement in performance \cite{borghans2008economics, paunonen2001big}, and longer-term outcomes in higher education and future earning \cite{heckman2006effects, deke2006valuing}. Recent studies in educational data mining have attempted to predict when an intervention can help students by distinguishing between productive and unproductive behavior using decision trees \cite{kai2018decision} and Recurrent Neural Networks (RNN) \cite{botelho2019developing}. The work by Kai et al. on ASSISTments used decision trees to identify when students are struggling and how to make students’ persistence more productive. They found that interleaved practice of different skills is more advantageous than blocked practice, where the opportunities to learn a given skill are massed one after another. Another study on ASSISTments by Botelho et al. used RNNs to detect stopout (low persistence) and wheel-spinning (unproductive persistence) early to intervene and prevent unproductivity. They found that these models have high AUC and are also able to learn a set of features that generalize to predict each other. In this paper, we apply clustering to discover patterns of productivity, persistence, and unsolicited hint usage in our tutor.

 \begin{figure*}
\centering
\includegraphics[width=\columnwidth]{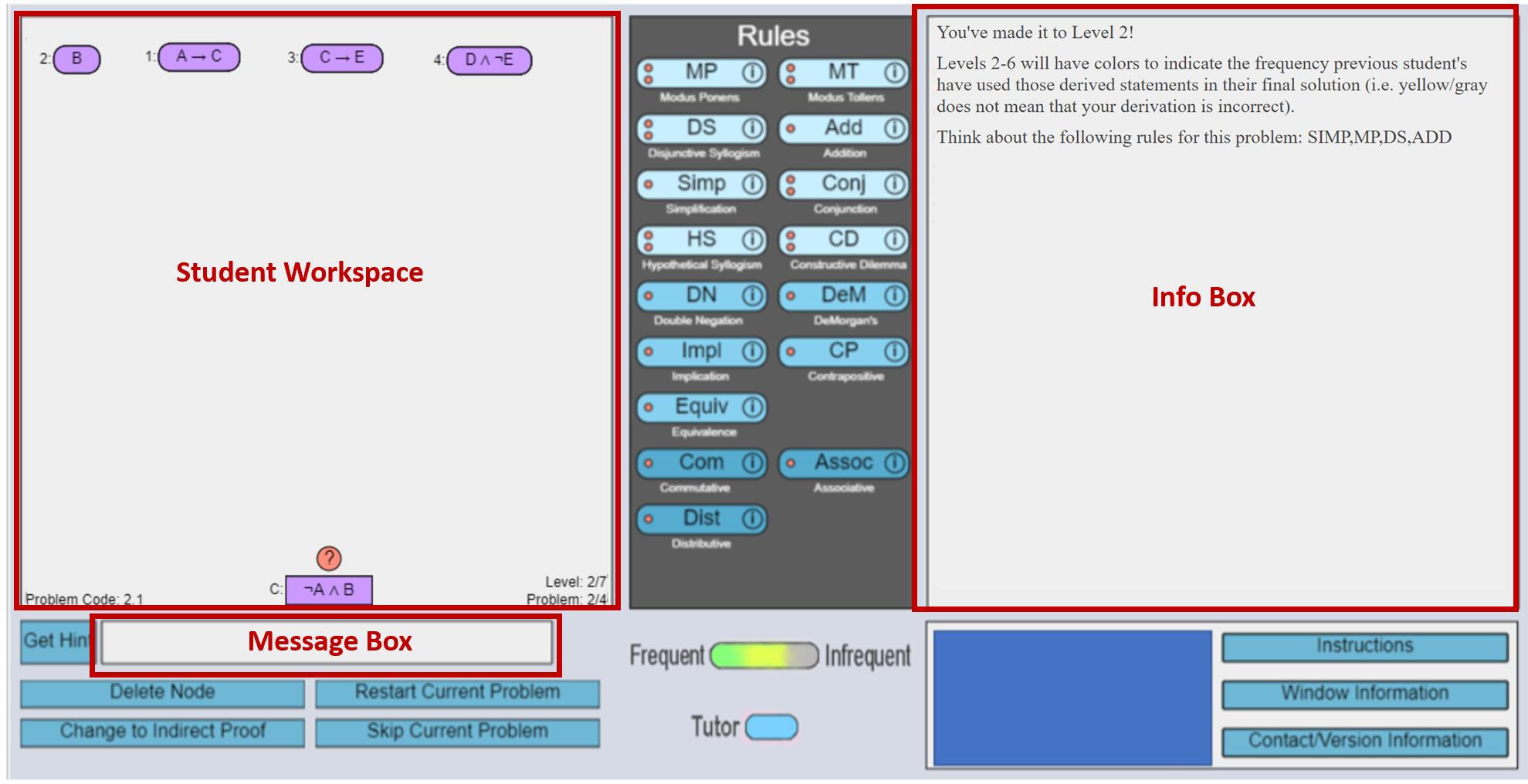}
\caption{Tutor's Interface: Student \textit{workspace} (left), rules (middle), info box (right), and the \textit{Hint button} and \textit{message box} (bottom-left)}
\label{fig:dt_ui}
\end{figure*}

\section{System Design}
\label{sys}
Deep Thought (DT, Figure \ref{fig:dt_ui}) is an intelligent tutor for solving open-ended multi-step propositional logic problems that has data-driven features including next-step hints \cite {stamper2008hint, barnes2010automatic}, as well as adaptive problem selection \cite{mostafavi2016data, mostafavi2015towards} and pedagogical policies for worked example presentation induced via reinforcement learning \cite{shen2016reinforcement, shen2018empirically, mostafavi2017evolution, ausin2019leveraging}. Figure \ref{fig:dt_ui} shows the current tutor interface: the left window is the \textit{workspace} where students construct solutions, the central window lists the domain rule buttons, and the right window provides instructions and information such as the rules that are meant to be practiced in the current problem. Each problem-solving statement is graphically represented as a \textit{node}. Deep Thought shows several problem-provided statements (that are meant to be used as existing or known facts) at the top of the workspace, and a conclusion to derive at the bottom. Students iteratively carry out problem-solving steps  by deriving new statements from old ones using domain rules. This is a typical procedure used across STEM domains to apply principles or rules to known information to derive new facts \cite{newell1972human}. For example, in physics, if we know values for mass ($m$) and acceleration ($a$), we can apply the rule $F=ma$ with those values to find force ($F$). In this paper, a problem-solving step consists of a new \textit{derived statement} and its \textit{justification}, where the justification includes specifying the domain rule and the source statements used to show that the new derived statement is true. In logic, problem-solving continues until the conclusion is the derived statement in a step that is justified.

\begin{figure*}
\centering
\caption{A sample solution of a training problem in Deep Thought}
\label{fig:demo}
\includegraphics[width=0.75\columnwidth]{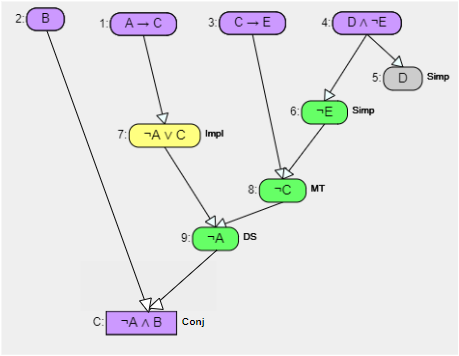} 
\end{figure*}

Figure \ref{fig:dt_ui} shows an example problem with three nodes 1-4 for the problem-provided statements (2: $B$, 1: $A \rightarrow C$, 3: $C \rightarrow E$), and 4: $D \land \neg E$ at the top of the workspace. The conclusion to be derived (C: $\neg A \land B$) is at the bottom, with a question mark indicating that it is not yet justified. Each problem solving step involves the same process: clicking on 1-2 source nodes and a rule button, and entering the new derived statement. The tutor verifies whether the source nodes and rule correctly justify the derived statement. Once a step is verified, a new node appears, colored based on how often the same node was necessary in previous student solutions to this problem, where green means frequent, yellow is infrequent, and gray is never. We call a node `necessary’ or `needed’ when its deletion would make a solution incomplete. These colorings give students an indication of whether they are on an optimal problem-solving path. 

We now walk through the student experience of solving the problem shown in \ref{fig:dt_ui} to obtain the solution shown in Figure \ref{fig:demo}. First, the student clicks on node 4 and rule Simp, and is asked to type the new derived statement, $D$.  The tutor verifies that Simp applied to node 4 is a correct justification, and draws node 5, labeled with Simp and an arrow from node 4 to 5. Node 5 is colored gray since it was never needed by previous students solving that same particular problem. Next, the student applies the same process to derive and justify node 6, which is green since it was frequently necessary in historical solutions. To derive node 7, the student clicks on node 1, and Impl rule, and types in the derived statement $\neg A \lor C$. After it is verified, node 7 appears, with the label Impl, and an arrow from node 1 to 7. The student then clicks ``Get Hint" to request a hint, and ``Try to derive $\neg C$" appears in the message box. Next, the student tries to follow the hint by selecting nodes 3 and 6 and the rule MP. The tutor detects this incorrect rule application, records the error in the data log, and provides an error-specific message, but since it was a mistake, no new node is created.  Since nodes 3 and 6 are still selected, the student clicks on the correct rule -- MT, and types in the derived statement $\neg C$. This process correctly \textit{justified} the hint content statement $\neg C$, so node 8 appears with MT with arrows from nodes 3 and 6.  The student similarly clicks on nodes 7 and 8, and rule DS to derive node 9. Finally, the student clicks on nodes 2 and 9, and rule Conj to derive the conclusion, and the tutor detects that the problem is complete. 

\subsection{Hints in Deep Thought}
Deep Thought uses the Hint Factory \cite{stamper2008hint} to generate hints, where the hint content depends only on the current problem solving state, a snapshot of a student problem-solving attempt. The Hint Factory \cite{stamper2008hint} works by treating problem-solving data from prior students as a Markov Decision Process and using value iteration to assign values to each state based on its distance from a valid observed solution. Then, the \textit{hint source} is set for a current student’s state by selecting the subsequent reachable state with the highest value. If the current state is not found, we rollback current student solution states until a matching state and its hint source are found. Finally, the Hint Factory-derived \textbf{hint content} is the \textit{newest derived statement} in the hint source state. Deep Thought inserts this derived statement, the hint content $HC$, into a template depending on the hint type, described below.

\begin{figure*}
\centering
\caption{Differences between Assertions and Messages while delivering a logic hint statement $A \rightarrow E$}
\label{fig:compare_ui}

\begin{subfigure}{.48\textwidth}
  \centering
  \includegraphics[width=\linewidth]{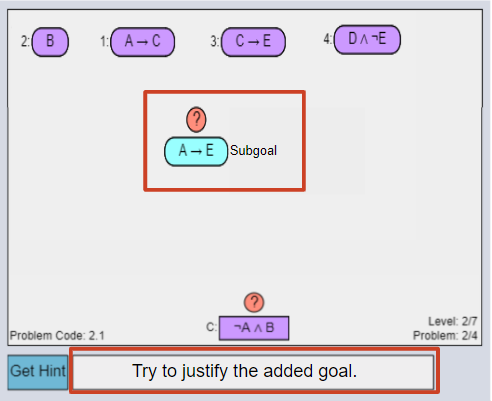}
  \caption{\textit{Assertion} is presented in the workspace, with the format of a student-derived step, and with a ``Subgoal" label}
 \label{fig:assertion}
\end{subfigure}\quad
\begin{subfigure}{.48\textwidth}
  \centering
  \includegraphics[width=\linewidth]{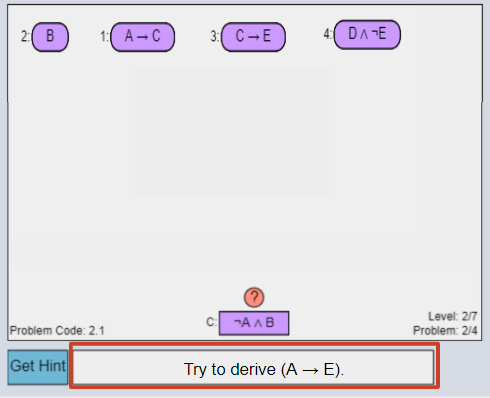}
  \caption{\textit{Message} hint is provided textually below the student workspace}
  \label{fig:message}
\end{subfigure}
\end{figure*}

In this study, there are three types of hints, including on-demand hint requests, and two types of unsolicited hints: Messages and Assertions. The content of on-demand and unsolicited hints is identical and no additional justification/derivation help is given. Students request \textit{on-demand} hints by clicking the ``Get Hint" button, and the system shows ``Try to derive $HC$," in the message box. Both Messages and Assertions are unsolicited hints, meaning that they are not requested by students. \textit{Messages} appear automatically after one minute of student inactivity, using the same Messages interface as on-demand hints. \textit{Assertions} appear automatically after about 40\% of steps. Since the mean student solution length in the training problems is 9 steps, this means that students are likely to encounter 3 - 4 hints per problem. The Assertions interface consists of 4 parts: (1) adding a new cyan-colored node containing the hint content $HC$ in the workspace, (2) labeling the node as a ``Subgoal,”, (3) including a question mark icon showing that the node is not yet justified, and (4) stating ``Try to justify the added goal" in the message box. Figure \ref{fig:compare_ui} shows the Messages and Assertions interfaces suggesting the same logic statement $A \rightarrow E$ in different formats. Students must \textit{explain} how the node is to be derived by \textit{justifying} it before they can use the hint content in their solutions. While this is not a typical verbal self-explanation, we argue that, by justifying the step, the student is demonstrating that they know what domain principle (rule) and prior statements can be used to \textit{explain} why the new derived statement is true. In the next section, we describe the design principles used to create the new Assertions interface.


\subsection{Assertions Design}
While there are many possible ways to encourage students to use hints more often \cite{conati2013understanding}, we hypothesize that our new Assertions interface changes would help all students notice and efficiently use unsolicited hints. As stated above, Assertions represent unsolicited, partially-worked example steps that appear in the workspace as shown in Figure \ref{fig:assertion}. In the remainder of this section, we discuss the design considerations that differentiate Assertions from Messages: contiguity, attention, expectation, and persuasion. All of these design changes were made to reduce students’ cognitive load \cite{sweller2008human}, and we hypothesized that these changes would reduce help avoidance as measured by increased hint usage. We further hypothesized that Assertions may have an aptitude-treatment interaction, with students with low prior knowledge benefiting more from the interface changes.

\begin{itemize}
\renewcommand{\labelitemi}{\textbullet }
\item \textbf{Contiguity and Attention}: Moreno, et al.’s \textit{spatial contiguity} principle for multimedia learning materials states that a graphic should not be physically separated from its explanatory text \cite{moreno1999cognitive, davies2013analysis}. Hegarty et al. showed that contiguity supports student memory and understanding \cite{hegarty1993constructing}. Butcher and Aleven showed that when interactive support was placed near a geometry diagram, student learning outcomes improved \cite{butcher2007integrating, butcher2013using}. Kardan and Conati in a controlled study on AI SPACE tailored the hint content, used hint highlighting and proximal hint placement to gain students attention towards unsolicited hints that improved learning for low prior knowledge students \cite{kardan2015providing}. In this work we use similar proximal hint placement for Assertions but provide the same content in both Assertions and Messages. We strategically place Assertion hints where the student needs them. Although the message box is close to the workspace, it may still be subject to `change blindness' \cite{healey2012attention}, where students paying \textit{attention} to nodes within the workspace may filter Messages out and simply not notice their appearance. Therefore, we provide Assertions in the workspace, where students have already focused their attention. Together, contiguity and attention are meant to help students notice the appearance of Assertion hints. 

 \item \textbf{Expectation}: Research by Summerfield explains that the speed of visual interpretation is optimized by leveraging past experiences to form expectations \cite{summerfield2009expectation}. Based on this principle, we design Assertions to leverage student expectations through an isomorphic visual format that may work together with reduced text to decrease cognitive load. First, the hint content $HC$ of Assertions appears in the same visual node format as student-derived statements, enabling students to visually interpret an Assertion hint faster. Second, Messages require students to read the text ``Try to derive $HC$” and determine that $HC$ is a statement that should appear on a graphical node. This additional cognitive processing may pose a barrier that some students may not overcome \cite{sweller2008human}, and this may be especially true for students with low prior knowledge \cite{kanfer1989motivation}. Therefore, formatting the Assertions hint content $HC$ as nodes may help students by leveraging visual expectation, or by reducing overall cognitive load \cite{sweller2008human}. 

\item \textbf{Persuasion}: Dillard suggests that user experiences can be enhanced by using persuasion \cite{dillard2013affect}. Cialdini has created six principles of influence, including reciprocity, commitment and consistency, liking, social proof, authority, and scarcity, that can be used to influence people’s behaviors \cite{cialdini2009influence}.  Assertions have two persuasive design aspects. First, we posit that adding Assertions directly to the workspace may make them seem required, leveraging the authority of the tutoring system itself. Assertion nodes are accompanied with a label ``subgoal" (Figure \ref{fig:assertion}) and the message ``Try to justify the added goal", persuasive and authoritative texts suggesting that justifying Assertions is just part of the tutor. The difference in the text accompanying Assertions and Messages is that an Assertion is called a ``goal” but message hints do use that terminology while providing hints. Second, Assertion nodes are also formatted with a question mark like the conclusion.  Formatting leverages both the visual expectation principle above, but also Cialdini’s consistency notion that people prefer to be consistent. Once they get used to following tutor instructions and justifying nodes that have question marks, Assertions can rely on people’s natural consistency that influences them to continue to make similar consistent choices. Previous studies on help-seeking and hint usage suggest that students have many different reasons for help avoidance, including their attitudes towards hints and their preference for autonomy \cite{price2017factors}. Persuasive design elements may circumvent these preferences by simply influencing students to do what is suggested. 
\end{itemize} 

\section{Method}
\label{usr}
Based on our foundational design principles and literature review, we propose the following three hypotheses: (H1) Assertions will increase the unsolicited hint usage for all students irrespective of their prior knowledge. (H2) Assertions will lead students with low prior knowledge to form shorter proofs faster in the posttest. (H3) Assertions will foster productive persistence among students with low prior knowledge.

\subsection{Participants}
The study was conducted with 122 participants at North Carolina State University, the top engineering university in the state, where Deep Thought was given as a homework assignment to a class of 312 undergraduate students in the College of Engineering majoring in Computer Science, Computer Engineering, or Electrical Engineering in a Fall 2018 discrete mathematics course. We do not have specific demographics of study participants, but the Fall 2018 College of Engineering demographics include 25.3\% women, 67.2\% white, 8.3\% Asian, 6.5\% Non-resident Alien, 0.3\% American Indian/Native American, 3.3\% Black/African American, 4.8\% Hispanic/Latinx, 4.83\% from two or more under-represented minorities, and 5.9\% with unknown race/ethnicity \footnote{More details can be found on Fall 2018 student demographics at NCSU at https://www.engr.ncsu.edu/ir/fast-facts/fall-2018-fast-facts/ The CSC 226 course is typically composed of about 60\% sophomores, 30\% juniors, 9\% seniors, and 1\% freshmen}.  

\subsection{Conditions}
\label{cond}
We used stratified sampling to split students based on their pretest performance, and then randomly assigned them to the conditions with \textit{Assertions} as the treatment, and \textit{Messages} as the control. The condition assignment resulted in $\mathit{N} = 73$ in Assertions, and $\mathit{N} = 49$ in Messages. The total number of participants who completed the study was 105 (61 in Assertions, 44 in Messages) but after removing logs with system errors, the dataset had 100 students with 57 in Assertions, and 43 in Messages.  We performed a $\chi^2$ test of independence to examine the impact of completion rate and system errors on the groups and found no significant differences among the two groups: $\mathit{\chi^2}(2, \mathit{N} = 122) = 1.88,  \mathit{p} = 0.91$. This implies that the group sizes were not significantly impacted by the tutor completion rate or logging errors. 

\subsection{Procedure}
The student procedure is as follows: The tutor provides students with practice solving logic problems, divided into four sections: introduction, pretest, training, and posttest. The introduction presents two worked examples to familiarize students with the tutor interface. Next, students solve two problems in a \textit{pretest}, which is used to determine students’ incoming competence. Students are assigned a condition based on their pretest performance. The pretest problems are designed to be easy and short, using a few straightforward rules, and this is reflected in their short optimal solution lengths ($\mathit{Mean} = 3.5$, $\mathit{SD} = 0.71$). Next, the tutor guides students through the \textit{training} section with five training levels with gradually increasing difficulty, and this is reflected in the average length of optimal solutions during training, with a mean optimal solution length of  $\mathit{Mean} = 4.99$ steps, ($\mathit{SD} = 1.32$). For each level, each student must solve \emph{four} training problems. Students may skip a maximum of three problems per level, with each skip taking students to easier problems. Students may also restart problems using the ``Restart" button below the workspace. In both conditions, students in the training levels may request on-demand hints and always receive immediate feedback on rule application errors (see section \ref{sys}). Students in the Messages (control) condition received unsolicited message hints upon one minute of inactivity.  Students in the Assertions (treatment) condition were given Assertions after about 40\% of their steps.The algorithm we use to provide Assertions uses two steps. In the first step, we decide at random whether the step should get a hint with 50\% probability. In the second step, we check for the constraints that assertion should not be given in more than two consecutive steps. This resulted in an actual assertion provision rate of 40\%. Note that both Messages and Assertions remain on the screen until a student justifies them\footnote{The tutor allows students to delete assertions but only two Assertions were deleted in the entire dataset, suggesting that students did not realize this was possible}. Further, only one unsolicited hint, regardless of interface, may be present at a time, and the hint content is not updated based on new student work. Finally, students take a more difficult \textit{posttest} with four problems, with longer optimal solution lengths compared to the other sections ($\mathit{Mean} = 7.25$, $\mathit{SD} = 1.89$). Students were given a week to complete the procedure. The average time students worked on the tutor was 3.4 hours, with a median of 2.6 hours and standard deviation of 2.8 hours.


\begin{figure*}
\centering
\caption{Example scenarios of Assertion hint $A \rightarrow E$ usage usage}
\begin{subfigure}{.45\textwidth}
  \centering
  \includegraphics[width=\linewidth]{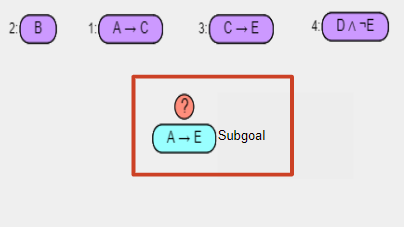}
  \caption{The Assertion $A \rightarrow E$ node appears in the student workspace; if it is never justified, it remains as-is}
  \label{fig:given}
\end{subfigure}%
\hspace{2em}
\begin{subfigure}{.49\textwidth}
  \centering
  \includegraphics[width=\linewidth]{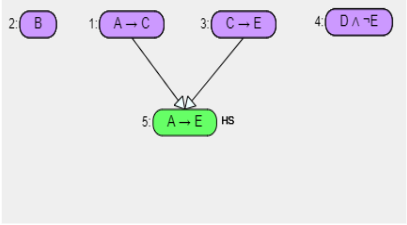}
  \caption{(The student has  \textit{justified} the hint by selecting nodes 1 and 3 and rule HS}
  \label{fig:justified}
\end{subfigure}

\vspace{2em}
\begin{subfigure}{.49\textwidth}
  \centering
  \includegraphics[width=\linewidth]{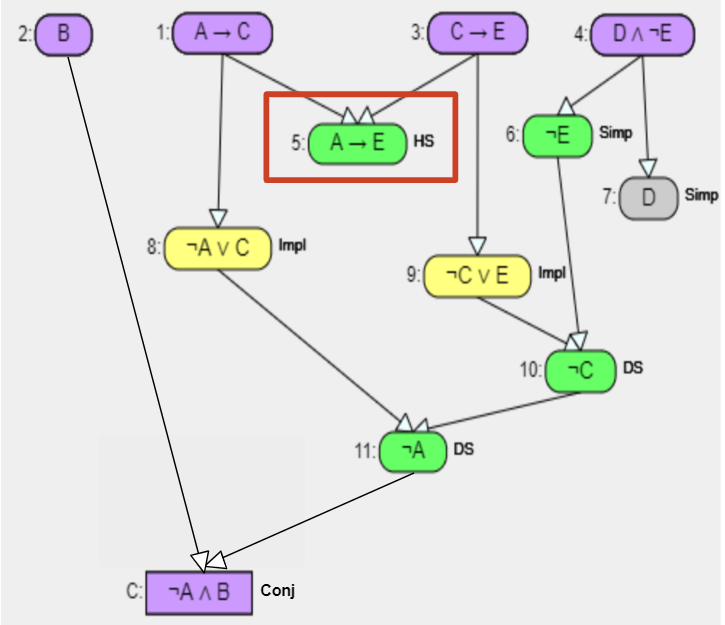}
  \caption{A student solution where hint $A \rightarrow E$ was justified but \textit{not needed}}
  \label{fig:example_just}
\end{subfigure}
\begin{subfigure}{.49\textwidth}
  \centering
  \includegraphics[width=\linewidth]{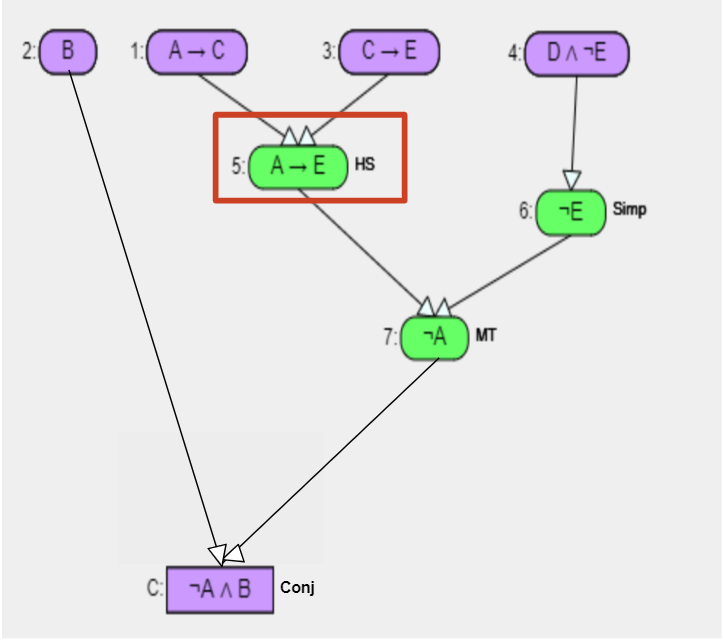}
  \caption{Another student solution where the hint was \textit{both justified and needed}}
  \label{fig:example_need}
\end{subfigure}
\label{fig:hint_usage}
\end{figure*}

\subsection{Hint Usage}
The motivation for using Assertions is to reduce students' reluctance towards using unsolicited help. We believe that increased attention will be paid to Assertions over Messages, and that this increased attention will lead to increased hint usage as prior studies have shown \cite{conati2013understanding}. Figure \ref{fig:hint_usage} illustrates two types of hint usage we can observe through tutor logs: hints justified and hints needed. A statement or hint is \textit{justified} when a student applies rules to existing statements to derive it. Figure \ref{fig:given} shows an Assertion suggesting $A \rightarrow E$. When a student selects nodes 1 and 3 and the rule \textit{HS} to derive the hint $A \rightarrow E$, the Assertion hint $A \rightarrow E$ is said to be justified, and it becomes a numbered node 5 as in Figure \ref{fig:justified}. The student may continue to solve the problem as in Figure  \ref{fig:example_just}, without ever having used node 5 to justify any other node. As in this case, whenever an Assertion was justified but could be deleted without making the solution incomplete, we say the Assertion was \textit{justified but not needed}. Another student may solve the problem as in Figure \ref{fig:example_need} where the same hint statement $A \rightarrow E$ is both justified and needed. If we remove node 5 from the solution, it becomes incomplete since nodes 7:$\neg A$ and C:$\neg a \land B$ could not be derived without it. 

We assume that if students justify a hint, they have paid attention to it. The \textit{Hint Justification Rate} (HJR) is defined as hints justified divided by the total given across the training problems. As in other multi-step open-ended problem domains, students may derive several statements that are not needed to solve a problem, making the solution longer than necessary. For a hint to be called \textit{needed}, students must first justify it, but must also figure out how they can use it to derive the conclusion. \textit{Hint Needed Rate (HNR)} is defined as hints needed divided by the total number of hints given across the training problems. We use unsolicited HJR to evaluate student attention towards unsolicited help, and unsolicited HNR to measure the influence of unsolicited hints on student problem solving.

\subsection{Performance Measures}
Our test performance measures include: solution length optimality, problem-solving time, and rule application accuracy. In open-ended domains, \textit{solution length}, i.e., the number of derived statements in a \textit{complete} solution, is a valuable performance metric as there is a vast diversity of possible student solution paths. Our aim with increasing unsolicited hint usage is to guide students to learn efficient problem-solving strategies from incorporating the partially worked example Assertion steps as necessary statements in their solutions. Since the posttest consists of four problems, we evaluate students based on their average solution length in the posttest, and shorter lengths are better\footnote{Note that solution length can only be calculated for complete solutions, and our data consists only of students who successfully completed the study by completing the mandatory pre- and post-test problems. \textit{N} = 5 (10\%) in Messages, and \textit{N} = 12 (16\%) in Assertions did not finish the tutor. A chi-square test shows no significant difference in the completion and non-completion group sizes between the two conditions ($\mathit{\chi^2}(1, \mathit{N} = 122) = 0.95,  \mathit{p} = 0.33$)}.

Problem solving time is also an important performance metric in open-ended domains. Similar to other studies \cite{kardan2015providing,tchetagni2002hierarchical}, we also assess students on the \textit{total time} they spend solving problems. In order to account for outliers, while calculating problem solving time, we cap each click-based interaction time to five minutes, i.e., if a student took more than five minutes to perform an interaction, we cap it to five \footnote{The $99^{th}$ percentile of interaction action time in Fall 2018 was 99.03s; 811 out of 260,750 interaction logs for 100 students in the study, had an action time greater than 5min}. A shorter problem solving time suggests better performance. We hypothesized that an increased usage of unsolicited hints, will help students learn to solve problems more quickly and with shorter solution lengths, and that these effects will be more pronounced for students with low prior knowledge.

Finally, \textit{Accuracy} is defined as the number of correct rule applications divided by the total number of applications. A higher accuracy value suggests better knowledge of how to apply domain rules. Since the tutor is designed to provide immediate feedback on incorrect rule applications without penalties, even within the pre- and post-tests (see section \ref{sys}), we do not hypothesize differences in the accuracy between the two conditions. We report accuracy for both conditions, however, for completeness.

\subsection{Prior Proficiency}
We hypothesize that an increase in the unsolicited hint usage significantly impacts the performance of students with low prior knowledge. Our prior work \cite{shen2016reinforcement} suggests that students with different incoming competencies can experience a treatment differently. To account for such aptitude-treatment interaction effects, we quantify prior knowledge by splitting the students into Low and High \textit{Prior Proficiency} groups using a normalized pretest performance score that combines the number of problem-solving steps, the average time spent on each step, and accuracy. The three performance measures are normalized separately and equally weighted in a combined score that is again normalized. Students with pretest performance $>$ 0.5 are classified as the High group, i.e., students with high prior proficiency, and students with lower pretest performance are classified as the Low group, i.e., students with low prior proficiency. We found an insignificant difference between the High and Low Prior Proficiency group sizes between the two Conditions ($\mathit{\chi^2}(1, \mathit{N} = 122) = 0.24,  \mathit{p} = 0.62$). This allows us to compare the students in the two Conditions within each Prior Proficiency group. 

\subsection{Effort and Persistence}
\label{effort}
Hypothesis H3 states that Assertions will encourage students with low prior proficiency to engage in productive persistence. Our definition of \emph{effort} is highly motivated by prior research. More specifically, Venture et al. defined a metric for students’ efforts as the amount of time spent on unsolved problems and they found that there was a significant correlation between the effort measured during the training and a self-report measure of persistence \cite{ventura2013relationship}. Later, in another study, they used this effort metric to measure persistence in an educational game that teaches Qualitative Physics \cite{ventura2013validity}. In our tutor, students can skip up to three problems per training level and thus we also measure the time students spent in these unsolved skipped problems as a measure of effort. Moreover, Dumdumaya et al. defined their effort metric as the number of reattempts made on a problem after a failed attempt predicted task persistence \cite{dumdumaya2018predicting}. In our tutor, this corresponds to the number of restarts on problems that students eventually solve. In the following, we separately track effort through two research-based measures: (1) time spent on unsolved (skipped) training problems, and (2) the number of restarts on solved training problems. For the purpose of this analysis, we define \emph{productive persistence} as persistent (high) efforts that result in higher posttest performance.

\section{Results}
After cleaning the data as described in Section \ref{cond}, an average of 2,483 interactions were logged and analyzed per student in our final sample of 100 students (with 57 in the Assertions condition, and 43 in Messages). We partitioned the students based on Prior Proficiency into Low (n = 41) and High (n = 59) groups. We then partitioned by Condition and Prior Proficiency resulting in 4 groups: Assertions-Low (n = 25), Assertions-High (n = 32), Messages-Low (n = 16), and Messages-High (n = 27). 

Before investigating any of our hypotheses, we first compared the number of on-demand hints between the two conditions to ensure that any differences between groups could not be explained by differences in on-demand hint requests. Similar to other tutors \cite{price2017factors, mathews2008does}, students in this study rarely request on-demand help irrespective of Condition or Prior Proficiency. We found no significant differences in the number of on-demand hint requests between conditions or by prior proficiency, with all conditions requesting, on average, less than one on-demand hint per problem.  Students in the Assertions condition requested few on-demand hints per problem, with Mean = 0.79 , SD = 3.92 (Assertions-Low group: Mean = 0.67, SD = 2.82, and Assertions-High group: Mean = 0.89, SD = 3.07). Students in the Messages condition similarly requested few on-demand hints per problem for the Messages group, with Mean = 0.55 , SD = 2.72 (Messages-Low: Mean = 0.46, SD = 2.36, and Messages-High: Mean = 0.59, SD = 2.43). The on-demand hint data was not normally distributed as tested by the Shapiro-Wilk’s test (Assertion: \textit{W} = 0.744, \textit{p} $<$ 0.001, Messages:  \textit{W} = 0.752, \textit{p} $<$ 0.001). So, a two-factor Aligned Ranks Transformation ANOVA \cite{wobbrock2011aligned} with the two factors as the Condition \{Assertion, Messages\} and Prior Proficiency \{Low, High\} on the number of on-demand hints shows no significant main effects (Condition: \textit{F}(1,100) = 0.132, \textit{p} = 0.718, Prior Proficiency: \textit{F}(1,100) = 1.075, \textit{p} = 0.302) or interaction (\textit{F}(1,100) = 0.006, \textit{p} = 0.940). Based on this analysis, the remaining analyses focus only on usage for unsolicited Assertion and Message hints.

\subsection{\textbf{H1}: Assertions increase the unsolicited hint usage for all students irrespective of their prior knowledge}
Table \ref{tab:hint_usage} shows the unsolicited hint metrics: \#Given (number of unsolicited hints given during training), HJR (Hint Justification Rate - proxy for attention paid), and HNR (Hint Needed Rate - hints' influence on problem-solving) for the two Conditions\footnote{HJR and HNR are the proportion of hints justified and needed respectively}. Since we have hint data that is not normally distributed\footnote{Shapiro-Wilk’s test on \textit{Unsolicited Hints Given} for the Assertions group: \textit{W} = 0.904, \textit{p} $<$ 0.001, and the Messages group: \textit{W} = 0.942, \textit{p} =  0.030; Shapiro-Wilk’s test on \textit{Unsolicited HJR} for the Assertions group:  \textit{W} = 0.887, \textit{p} $<$ 0.001, the Messages group: \textit{W} = 0.959, \textit{p} $<$ 0.001; and  Shapiro-Wilk’s test on \textit{Unsolicited HNR} for the Assertions group:  \textit{W} = 0.904, \textit{p} $<$ 0.001, and the Messages group:  \textit{W} =0.945, \textit{p} $<$ 0.001}, we performed a two-way Aligned Ranks Transformation ANOVA \cite{wobbrock2011aligned} on each of the unsolicited hint metrics with the two factors as Condition \{Assertions, Messages\}, and Prior Proficiency \{Low, High\}.

\begin{table}
\centering
\caption{Comparison of unsolicited hint metrics between the two conditions, where a two-way Aligned Ranks Transformation ANOVA for each metric shows only a main effect of Condition (\textit{p} $<$ 0.001*)}
\begin{tabular}{r|cc}
\hline
Unsolicited
    Hint Metric   & Assertions    & Messages        \\ \hline
Hints Given in Training       & 48.82 (9.85)* & 32.74 (10.64)  \\\rowcolor[HTML]{EFEFEF}
Hint Justification Rate (HJR) & 0.93 (0.07)*  & 0.63 (0.18)    \\ 
Hint Needed Rate (HNR)        & 0.82 (0.09)* & 0.62 (0.17)   
\end{tabular}
\label{tab:hint_usage}
\end{table}

We applied a two-way Aligned Ranks Transformation ANOVA on the unsolicited hint metrics of \#Given, HJR and HNR as described above. For the \#Given metric, we found a significant main effect of \textit{Condition} (\textit{F}(1,100) = 40.26, \textit{p} $<$ 0.001). While we expected this result because of the system design, we will discuss this further in the next section. For both hint usage metrics HJR and HNR, we observed a significant main effect of \textit{Condition} (HJR: \textit{F}(1,100) = 191.10, \textit{p} $<$ 0.001, and HNR: \textit{F}(1,100) = 62.30, \textit{p} $<$ 0.001). The main effect of Prior Proficiency and the interaction effect were not significant for either HJR or HNR. As we hypothesized, the Assertions groups used a significantly higher proportion of unsolicited hints, both by justifying (HJR) and needing (HNR) more unsolicited hints than the Messages group, as shown in Table \ref{tab:hint_usage}.

We did not observe a significant interaction between Condition and Prior Proficiency for any of the unsolicited hint metrics: \#Given: \textit{F}(1,100) = 0.008, \textit{p} = 0.929, HJR: \textit{F}(1,100) = 0.221, \textit{p} = 0.639, and HNR: \textit{F}(1,100) = 0.009, \textit{p} = 0.924. The distribution parameters for each of the unsolicited hint metrics per Prior Proficiency group are provided in Appendix \ref{appendix_hint_usage}. There was only a main effect of the Condition as shown above. This shows that the Assertions had a significant impact on unsolicited hint usage for all students, regardless of incoming proficiency, confirming hypothesis H1.

\subsection{\textbf{H2}: Assertions will lead students with low prior knowledge to form shorter proofs faster in the posttest}

Since all performance data were normal, we performed t-tests to compare conditions. A t-test on the average pretest solution length between the Assertions (Mean = 7.54 nodes, SD = 1.87 nodes) and the Messages (Mean = 7.64, SD = 2.21) conditions, showing no significant difference (\textit{t}(99) = 0.791, \textit{p} = 0.215). We also observed insignificant differences in the pretest problem-solving time (\textit{t}(99) = 0.683, \textit{p} = 0.248) using a t-test between the Assertions (Mean = 27.16 min, SD = 8.29 min) and the Messages (Mean = 25.89 min, SD = 10.23 min) conditions. While the H2 hypothesis does not predict differences in accuracy between conditions, students were assigned a condition based on their pretest performance, which includes rule application accuracy, so we compare it here. A t-test on the pretest rule accuracy between the Assertions (Mean = 0.52, SD = 0.16) and Messages (Mean = 0.52, SD = 0.14) conditions shows no significant difference (\textit{t}(99) = 0.111, \textit{p} = 0.455).

\begin{table*}[]
\centering
\caption{Comparison of \textbf{\textit{Posttest Performance metrics}} between the two conditions within each \textbf{\textit{Prior Proficiency group}} - Average Solution Length (\textit{p} = 0.033) and Total time (\textit{p} = 0.008) are significantly different between the Assertions-Low and the Messages-Low groups}
\begin{tabular}{c|cc|cc}
\hline
\multirow{3}{*}{\begin{tabular}[c]{@{}c@{}}Prior\\ Profi- \\ciency\end{tabular}} & \multicolumn{2}{c|}{Avg. Sol. Length (\#nodes)} & \multicolumn{2}{c}{Total Time (min)}  \\ \cline{2-5} 
                                                                             & Assertions          & Messages            & Assertions        & Messages          \\ 
                                                                             & Mean (SD)           & Mean (SD)           & Mean (SD)         & Mean (SD)      \\\rowcolor[HTML]{EFEFEF} \hline

\color{blue}Low & \color{blue}13.33 (1.10)*        & \color{blue}15.09 (1.95)        & \color{blue}36.03 (12.81)*     & \color{blue}52.94 (18.19)     \\  
High                                                                         & 14.36 (1.57)        & 14.49 (1.38)        & 41.73 (17.12)     & 43.27 (19.09)      \\ \rowcolor[HTML]{EFEFEF} \bottomrule
All              & 13.92 (1.44)        & 14.46 (1.69)        &38.13 (14.95)    & 46.99 (17.65)   \\ 
\end{tabular}
\label{tab:posttest}
\end{table*}

 \begin{figure*}
\centering
\includegraphics[width=0.7\columnwidth]{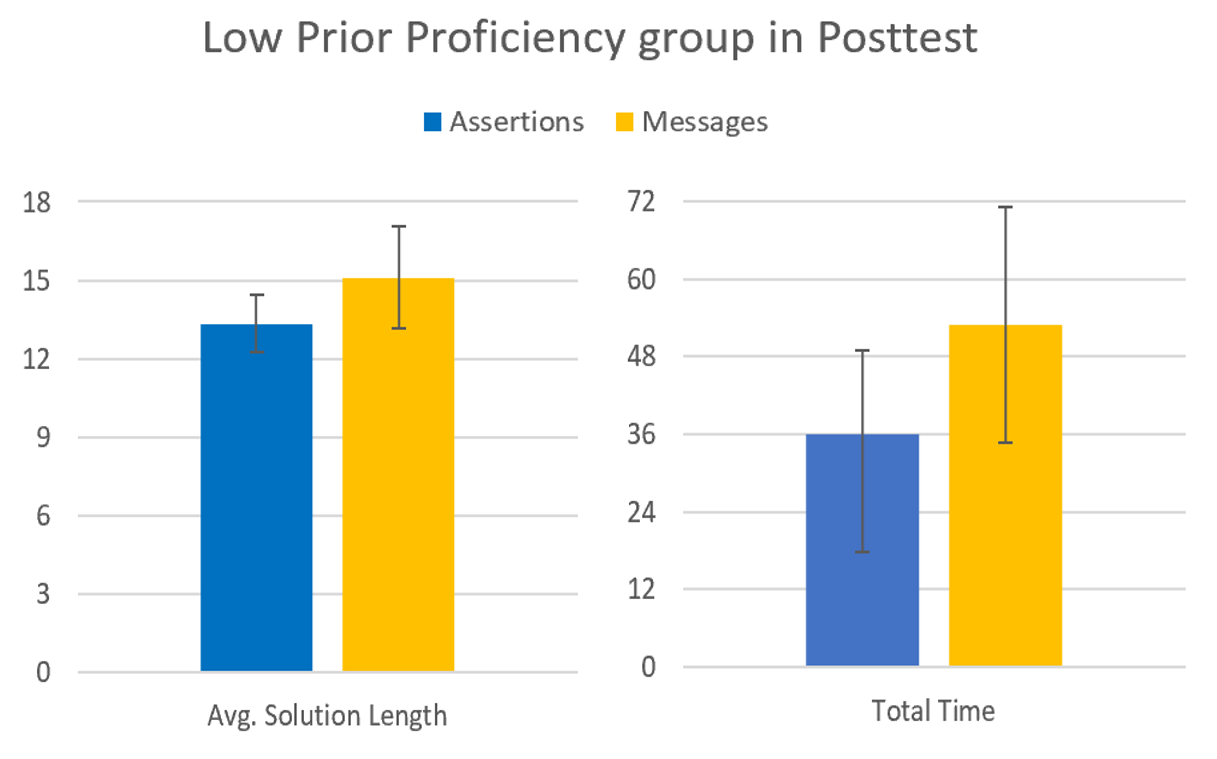}
\caption{Tukey’s HSD shows that the Assertions-Low performed significantly better in posttest than the Message-Low group in average solution length (\textit{p} = 0.033) and total time (\textit{p} = 0.008)}
\label{fig:posttest_graph}
\end{figure*}

As mentioned earlier, hypothesis H2 is based on the reasoning that Assertions may guide students towards optimal strategies, which can lead students with low prior proficiency to form shorter solutions in less time. We examined the correlation between the dependent variables (average solution length and total time) to assess their overlap, both for the entire population and for the low prior proficiency group. We did not observe a significant correlation between the average posttest solution length and posttest time for the entire population: \textit{Corr} = 0.050, \textit{p} = 0.615 or for the Low Prior Proficiency group: \textit{Corr} = 0.015, \textit{p} = 0.916.

Table \ref{tab:posttest} shows the posttest performance of the two Conditions \{Assertions, Messages\} disaggregated for the  \textit{Low}, and \textit{High} Prior Proficiency groups in the first two rows, and for  \textit{All} students as a summary in the bottom row. To investigate our H2 hypothesis, we performed a two-way ANCOVA on average solution length and total time, with the Condition \{Assertions, Messages\} and Prior Proficiency \{Low, High\} as the two factors, and the respective pretest performance metric as the covariate. For average solution length, we observed a significant interaction between the Condition and Prior Proficiency (\textit{F}(1,100) = 4.983, \textit{p} = 0.027). Neither main effect for Condition or Prior Proficiency were significant. We then performed the pairwise Tukey's Honest Significant (HSD) test for multiple comparisons and found a significant difference (\textit{p} = 0.033) between the \textit{Assertions-Low} and \textit{Messages-Low} groups, showing that the Assertions-Low group formed significantly shorter proofs on the posttest than the Messages-Low group. 

A two-factor ANCOVA on posttest total time as described above shows a significant interaction between the Condition and Prior Proficiency (\textit{F}(1,100) = 6.236, \textit{p} = 0.014), and a significant main effect of the Condition (\textit{F}(1,100) = 6.913 \textit{p} = 0.010). The main effect of Prior Proficiency was not significant. A pairwise Tukey's Honest Significant (HSD) test for multiple comparisons on the total posttest time shows a significant difference (\textit{p} = 0.008) between \textit{Assertions-Low} and \textit{Messages-Low} groups. The Assertions-Low group spent significantly less time on the posttest than the Messages-Low group. Figure \ref{fig:posttest_graph} summarizes the differences between the Assertions-Low and Messages-Low groups in their posttest performance. Together with the results above, the Assertions-Low group had significantly better posttest solution length and time than the Messages-Low group, confirming our H2 aptitude-treatment interaction hypothesis for posttest performance. While we did not hypothesize improvements in posttest accuracy, we provide these results in Appendix \ref{apx_accuracy} for completeness.


\begin{table*}[]
\centering
\caption{Correlation between \textbf{\textit{average posttest solution length}} and unsolicited hint metrics for the entire population, and split by low and high prior proficiency groups}
\begin{tabular}{r|cccccc}
\multicolumn{1}{c|}{\multirow{2}{*}{\begin{tabular}[c]{@{}c@{}}Posttest \\ Solution Length\\ with\end{tabular}}} & \multicolumn{2}{c}{\begin{tabular}[c]{@{}c@{}}Entire \\ Population\\ N = 100\end{tabular}} & \multicolumn{2}{c}{\begin{tabular}[c]{@{}c@{}}Low Prior \\ Proficiency\\ N = 41\end{tabular}} & \multicolumn{2}{c}{\begin{tabular}[c]{@{}c@{}}High Prior \\ Proficiency\\ N = 59\end{tabular}} \\ \cline{2-7} 
\multicolumn{1}{c|}{} & \multicolumn{1}{c}{Corr} & \multicolumn{1}{c}{\textit{p}} & \multicolumn{1}{c}{Corr} & \multicolumn{1}{c}{\textit{p}} & \multicolumn{1}{c}{Corr} & \multicolumn{1}{c}{\textit{p}} \\ \rowcolor[HTML]{EFEFEF}  \hline
\#Given (Frequency) & -0.13 & 0.18 & -0.10 & 0.26 & -0.03 & 0.83 \\ 
HJR (Attention) & -0.25 & 0.01* & \color{blue}-0.37 & \color{blue} $<$ 0.001* & -0.07 & 0.65  \\\rowcolor[HTML]{EFEFEF} 
HNR (Influence) & -0.02 & 0.87 & \color{blue}-0.48 & \color{blue}$<$0.001* & 0.12 & 0.42 
\end{tabular}
\label{tab:corr_len}

\vspace{3em}

\centering
\caption{Correlation between \textbf{\textit{total posttest time}} and unsolicited hint metrics for the entire population, and split by low and high prior proficiency groups}
\begin{tabular}{r|cccccc}
\multicolumn{1}{c|}{\multirow{2}{*}{\begin{tabular}[c]{@{}c@{}}Posttest \\ Total Time\\ with\end{tabular}}} & \multicolumn{2}{c}{\begin{tabular}[c]{@{}c@{}}Entire \\ Population\\ N = 100\end{tabular}} & \multicolumn{2}{c}{\begin{tabular}[c]{@{}c@{}}Low Prior \\ Proficiency\\ N = 41\end{tabular}} & \multicolumn{2}{c}{\begin{tabular}[c]{@{}c@{}}High Prior \\ Proficiency\\ N = 59\end{tabular}} \\ \cline{2-7} 
\multicolumn{1}{c|}{} & \multicolumn{1}{c}{Corr} & \multicolumn{1}{c}{\textit{p}} & \multicolumn{1}{c}{Corr} & \multicolumn{1}{c}{\textit{p}} & \multicolumn{1}{c}{Corr} & \multicolumn{1}{c}{\textit{p}} \\ \rowcolor[HTML]{EFEFEF}  \hline
\#Given (Frequency) & -0.02 & 0.88 & -0.06 & 0.68 & -0.17 & 0.23 \\ 
HJR (Attention) & -0.28 & $<$0.01* & \color{blue}-0.40 & \color{blue} $<$ 0.001* & -0.25 & 0.07  \\\rowcolor[HTML]{EFEFEF} 
HNR (Influence) & -0.27 & $<$0.01* & \color{blue}-0.36 & \color{blue}$<$0.001* & -0.20 & 0.16 \label{tab:corr_time}
\end{tabular}
\end{table*}

Next, we investigated the correlation between average posttest solution length with the unsolicited hint metrics. First, the top row of Table \ref{tab:corr_len} shows that the number of unsolicited hints given does not correlate to posttest solution length, suggesting that differences in posttest solution lengths between conditions cannot be attributed to the frequency of unsolicited hints. However, both HJR (second row) and HNR (third row) are significantly and negatively correlated to posttest solution length for students with Low Prior Proficiency (HJR: \textit{p} $<$ 0.001, HNR: \textit{p} $<$ 0.001), with a stronger correlation to posttest solution length for HNR than HJR\footnote{We did not test for the significance in the difference between the two correlation coefficients because the samples are not independent. Hints Needed are a subset of Hints Justified}. This suggests that students with low prior knowledge learn more from the hints needed, rather than the ones they only justified (see Figure \ref{fig:hint_usage} differentiating hints justified and needed). A justified, but not needed, hint suggests that a student could determine how to derive the unsolicited hint content, but not \textit{how to use it}. It is reasonable that lower prior proficiency students who were able to include the unsolicited hints as necessary components of their proof solutions were more likely to learn more optimal, shorter problem-solving strategies. We also observed an insignificant but positive correlation between average posttest solution length and HNR for the High prior knowledge group. While small and not significant, this inverted effect may indicate another aspect of aptitude treatment interaction, where high prior proficiency students may potentially learn less if they take too much advantage of unsolicited hints. This result suggests that it may be preferable to build a more adaptive method to determine when to present unsolicited hints to students with high prior proficiency.

Table \ref{tab:corr_time} shows the correlation between posttest time and the unsolicited hint metrics. First, Table \ref{tab:corr_time} shows that the number of unsolicited hints given (top row) does not correlate to posttest time, suggesting that differences in posttest time between conditions cannot be attributed to the frequency of unsolicited hints. While HJR (second row) and HNR (third row) are significantly correlated to the posttest time for the entire population, the Pearson's Correlation Coefficient is less than 0.3, suggesting small coverage. However, students with Low Prior Proficiency have a significant correlation (that is also greater than 0.3) between posttest time and unsolicited hint usage metrics HJR and HNR.
  
Table \ref{tab:hint_usage} shows that, over the entire population, significantly more (\textit{p} $<$ 0.001) unsolicited hints were given on average (\#Given) in the Assertions condition (39.78 \% of total steps on average) than in Messages (26.93 \% of total steps on average). We observed a significant main effect of \textit{Condition} on the number of unsolicited hints given. Neither the main effect of Prior Proficiency nor the interaction effect were significant. It would be reasonable to expect that the frequency of hints might impact posttest performance. However, our correlation analysis shows that the significantly higher number of unsolicited hints given in the Assertions condition did not correlate with posttest performance for either solution length or time. Instead, the significant negative correlations between posttest length and time, and Hints Needed Rate for all students with low prior knowledge suggests that students in the Low group learned from using the unsolicited hints to achieve problem conclusions. These needed hints provided insight into efficient problem solving, by showing students optimal problem-solving steps. As shown in Table \ref{tab:hint_usage} above, students in the Assertions condition had higher HNR than students in the Messages condition. Therefore, our results confirm hypothesis H2 that there would be an aptitude-treatment interaction effect where Assertions helped students with low prior proficiency learn to construct more optimal (shorter) solutions more quickly on the posttest.

\subsection{\textbf{H3}: Assertions foster productive persistence among students with low prior knowledge}
We hypothesized that increased usage of unsolicited hints in the form of Assertions will lead students with low prior proficiency to exert persistent effort in training, and this persistence will be productive (i.e., improved posttest performance). We clustered students on five features including: two productivity measures (posttest solution length and time, where lower is better), two effort measures including time spent on unsolved (skipped) problems and the number of restarts, and unsolicited hint usage as measured by HJR. We used Hint Justification Rate (HJR) instead of Hint Needed Rate (HNR) since the hints needed cannot be determined for unsolved problems. The clustering analysis provides a deeper understanding of student behavior patterns involving productivity, effort, and proactive hint usage. An ANOVA on the effort metrics would not have helped us understand how student effort varies in tandem with both productivity and hint usage. Therefore, the cluster analysis is more geared towards answering H3 than an ANOVA.

\begin{table}
\caption{Selecting the number of clusters based on three cluster quality indices}
\centering
\begin{tabular}{c|ccc}
\#Clusters &  Silhouette  & \begin{tabular}[c]{@{}l@{}}Davies-\\Bouldin\end{tabular} & \begin{tabular}[c]{@{}l@{}}Calinski-\\Harabasz~\end{tabular}  \\ \rowcolor[HTML]{EFEFEF} 
\hline
2  & 0.49  & 0.63 & 122.36\\
3 & \color{blue}\textbf{0.55}  & \color{blue}\textbf{0.51} & 234.91 \\\rowcolor[HTML]{EFEFEF}
4 & 0.50          & 0.57 & 271.28   \\
5 & 0.46         & 0.61 & \color{blue}\textbf{285.74}                                            
\end{tabular}
\label{tab:select_clusters}
\end{table}

We performed cluster analysis using Hierarchical clustering with Ward’s method on standardized features. We selected the number of clusters using majority vote across three indices: Silhouette and Calinski-Harabasz, which both maximize inter-cluster similarity and minimize intra-cluster similarity (overall higher values are better), and the Davies-Bouldin Index, which prefers minimal intra-cluster similarity (overall lower values are better). Table \ref{tab:select_clusters} shows that using three clusters yields the best quality clusters.

Table \ref{tab:centroids} shows the centroids of the three clusters. We used the \textit{class average (CA)}, i.e., average over the entire population to assess the clusters on each measure. The following order was observed for each feature used in the clustering analysis: (Note that lower posttest time and solution lengths are better)
\begin{itemize}
\renewcommand{\labelitemi}{\textbullet }
\item \textbf{Posttest Time (min)}: \#1 $<$ \#2 $<$ CA (42.81) $<$ \#3 
\item \textbf{Posttest Sol. Length}: \#1 $<$ \#2 $<$ CA (14.01) $<$ \#3 
\item \textbf{Unsolved Problem Time (min)}: \#1 $>$ CA (5.16) $>$ \#2 $>$ \#3 
\item \textbf{Restarts}: \#1 $>$ CA (2.44) $>$\#2 $>$ \#3
\item \textbf{Hint Justification Rate (HJR)}: \#1 $>$ \#2 $>$ CA (0.80) $>$  \#3
\end{itemize}


\begin{table}[]
\caption{Centroids (Mean) of the three clusters using Hierarchical Clustering with the Ward's method}
\centering
\begin{adjustwidth}{-0.5cm}{}
\begin{tabular}{c|l|cc|ccc}
\multirow{2}{*}{\begin{tabular}[c]{@{}c@{}} \\Clus-\\ter \\ No.\end{tabular}} & \multicolumn{1}{c|}{\multirow{2}{*}{\begin{tabular}[c]{@{}c@{}} \\Cluster \\ Label\end{tabular}}}   & \multicolumn{2}{c|}{Posttest} & \multicolumn{3}{c}{Training}             \\ \cline{3-7}
                                                                        & \multicolumn{1}{c|}{}                                                                          & \begin{tabular}[c]{@{}c@{}}Total\\ Time \\(min)\end{tabular} & \begin{tabular}[c]{@{}c@{}}Avg. \\ Sol. \\ Length \\(\#nodes)\end{tabular} &  \begin{tabular}[c]{@{}c@{}}Unsolved\\  Problem \\ Time\\ (min)\end{tabular} & \begin{tabular}[c]{@{}c@{}}Re-\\starts\end{tabular} &HJR   \\ \rowcolor[HTML]{EFEFEF} \hline
\#1                                                                       & \begin{tabular}[c]{@{}l@{}}Productive - High Effort - High HJR\end{tabular}     & 30.11   & 13.08 & 19.62 & 4.58  & 0.88 \\
\#2                                                                       & \begin{tabular}[c]{@{}l@{}}Productive - Low Effort - High HJR\end{tabular}   & 38.51    & 13.98  & 3.79    & 1.63 & 0.83                                                         \\\rowcolor[HTML]{EFEFEF}
\#3                                                                       & \begin{tabular}[c]{@{}l@{}}Unproductive - Low Effort - Low HJR\end{tabular}         & 59.19  & 14.75 & 0.81     & 1.06   & 0.50                                                        
\end{tabular}
\end{adjustwidth}   
\label{tab:centroids}
\end{table}

\begin{figure}
\centering
\includegraphics[width=0.75\columnwidth]{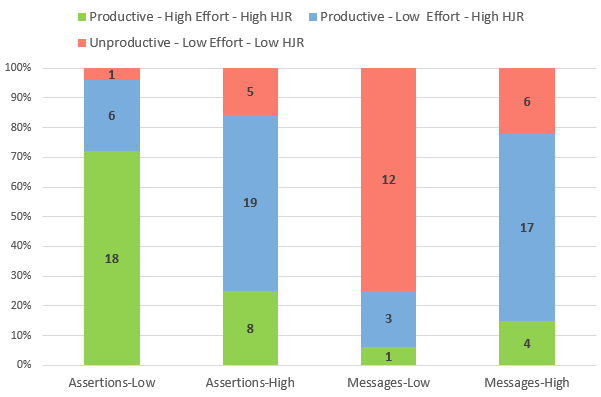}
\caption{Profile of the three Clusters based on the Condition and Prior Proficiency}
\label{fig:persist}
\end{figure}

As shown in Table \ref{tab:centroids}, Cluster \# 1 shows the highest productivity in terms of posttest performance, with the highest effort averages, and HJR (with all five features better than the class average). In the following, we refer to  this cluster as \textit{\textbf{Productive - High Effort- High HJR}}. Cluster \#2's posttest performance and HJR measures are better than the class average but their average effort on both time on unsolved problems and number of restarts are lower than the class average. This cluster was productive without needing to exert a high effort on unsolved problems or restarting problems during training. So, we call this cluster \textit{\textbf{Productive - Low Effort - High HJR}}. Interestingly, a lot of the High Prior Proficiency students ended up in low effort but did no better than Assertions-Low  group on the posttest.  Lastly, cluster \#3 shows the worst posttest time and solution length (both higher than the class average), the lowest effort on time on unsolved problems and number of restarts (both lower than the class averages), and the lowest HJR (lower than the class average), so we label this cluster \textit{\textbf{Unproductive - Low Effort- Low HJR}}.

We then profiled each cluster based on the pairs of the Condition and Prior Proficiency as shown in Figure \ref{fig:persist}. Interestingly, the majority of the Assertions-Low group students are in the \textit{Productive - High Effort- High HJR} cluster, and the majority of the Messages-Low group students are in the \textit{Unproductive - Low Effort- Low HJR} cluster. Most of the students in the Assertions-High and the Messages-High groups are in the \textit{Productive - Low Effort- High HJR} cluster. Since we are interested in the Low Prior Proficiency group, we performed a chi-square test to compare the distribution of the Assertions-Low and Messages-Low students in the three clusters and found a significant difference ($\mathit{\chi^2}(1, \mathit{N} = 41) = 24.73,  \mathit{p} < 0.001$). The majority of the Assertions-Low students show persistent effort as they are in the \textit{Productive - High Effort- High HJR} cluster with the highest effort and unsolicited hint usage in training with productive posttest results, and this confirms our H3 hypothesis.

\section{Discussion}
\subsection{\textbf{H1}: Assertions increase the unsolicited hint usage for all students irrespective of their prior knowledge}
The hints in our tutor suggest the most optimal next-step statement to derive for any given student problem-solving state. Similar to other tutors \cite{price2017factors, mathews2008does}, in this study, we found that students rarely request on-demand help irrespective of condition. However, our results suggest that the difference in unsolicited hint usage between Messages and Assertions can be attributed to presentation alone. We found that Assertions, specifically designed using the principles of contiguity, attention, expectation, and persuasion, significantly increased both the attention students pay to unsolicited hints (HJR), and their influence on students’ solutions (HNR) regardless of the students’ prior knowledge. Conati and Manske suggested in \cite{conati2009evaluating} that students pay more \textit{attention} to simpler hints. Assertions provide high immediacy (making the hint content immediately usable, \cite{bakke2014immediacy}) since they leverage both spatial \textit{contiguity} \cite{moreno1999cognitive} by placing information right where it is needed and visual \textit{expectation} \cite{summerfield2009expectation} by formatting hints to make them more intuitive to follow. Studies have also found students’ attitude towards unsolicited hints to be an important factor in help avoidance \cite{conati2013understanding, price2017factors}. Persuasive factors like increasing perceived authority \cite{cialdini2009influence} through formatting and language can make justifying Assertions seem to be required. Our results show that an unsolicited hint interface that combines persuasion, making hint usage seem required, with high immediacy, making it easy to see and do, can help overcome barriers to hint usage.

\subsection{\textbf{H2}: Assertions will lead to students with low prior knowledge to form shorter proofs faster in the posttest}
Several studies have found ATI effects surrounding hint usage where students with low prior knowledge or proficiency benefit more from interventions \cite{arroyo2001analyzing,murray2006comparison,kardan2015providing}. In particular, Kardan and Conati \cite{kardan2015providing} found that attention to hints affected student performance in a tutor for teaching constraint satisfaction problems, and students with low prior knowledge experienced more pronounced effects from an adaptive hint design intervention. While their intervention dealt with both an unsolicited hint interface (highlights to direct attention) and scaffolding (incremental textual hints), our study focuses only on the interface of unsolicited hints. Our ANCOVA results showed a significant aptitude-treatment interaction between Prior Proficiency \{High, Low\} and Condition \{Assertions, Messages\}. Using Tukey's HSD tests, we inferred that the Assertions-Low group outperformed the Messages-Low group in posttest solution length and time. We also observed a significant correlation of the posttest solution length and time with hint needed rate (HNR) for the Low Prior Proficiency group, suggesting that using more unsolicited hints as necessary components of their proofs helped this group learn better strategies. However, no such relations were observed for the High Prior Proficiency group. This suggests that adapting hint timing of Assertions based on proficiency may improve student performance as in other ITSs \cite{timms2007using, wood1999help, villesseche2018enhancing}. 

\subsection{\textbf{H3}: Assertions foster productive persistence among students with low prior knowledge}
Persistent effort is said to be productive when it is accompanied by an improvement in posttest performance \cite{borghans2008economics}. Assertions are designed to encourage students to follow unsolicited hints that direct students toward optimal problem-solving strategies. Results from our empirical study support the notion that Assertions promote productive persistence. Our cluster analysis showed that the majority of the Assertions-Low group exerted more effort (high persistence) during training, justified a higher proportion of unsolicited hints, and performed better on the posttest than the class average. We also saw a higher proportion of the Messages-Low students in the cluster that exerted less effort (low persistence) in the training, justified a lower proportion of unsolicited hints, and performed worse on the posttest than the class average. Interestingly, while most of the Assertions-Low group spent more time on unsolved problems in training, they took a significantly shorter time on the posttest while creating shorter posttest solutions, suggesting that the Assertions promoted productive persistence (i.e. time well spent) among students with low prior proficiency.

\subsection{Assertions - a new genre of hints}
Overall, this study showcases the importance of effective delivery for unsolicited hints, and a new genre of hints that we call Assertions. We believe that providing unsolicited hints as partially worked steps reduced the cognitive load required for learning from them. Further, increasing spatial contiguity improved students’ attention towards hints, and the isomorphic format may have made it easier for them to understand and use them in their solutions. We observed that Assertions led students with low prior knowledge to exert more productive persistence in training that resulted in better posttest performance, where they formed significantly shorter, more optimal, solutions in significantly less time than their peers in the control condition who only received Message hints. Assertions provide students with additional problem-solving resources that can enable them to learn through the process of self-explaining (justifying) expert steps. We believe that Assertions may be particularly helpful in multi-step domains, where providing students with partially-worked steps, right next to where they are needed, periodically, and in the same format as other problem-solving steps, could lead students to do more self-explanation (through justifying or completing the partially-worked steps) and by circumventing help avoidance.  

A limitation of this work is the difference in the timing of Assertions and Messages, which could have impacted the results. While the hint frequency correlation analysis showed that students’ posttest performance was not impacted by the \textit{number} of unsolicited hints given, we recognize that the hint timing may have had an impact on students. This limitation arises from the fact that we are modifying a real adaptive system to achieve practical improvements. These two types of hints were designed for different purposes. Messages were intended to help someone who was struggling but forgot about the help feature. Assertions were intended to be proactive for students who wouldn’t ask for help no matter what.Assertions were designed to address the problem that we observed, that Messages were not helping enough people improve their performance or learning.

\section{Conclusion and Future Work}
In this study, we investigated the impact of Assertions, a new genre of unsolicited hints, on the hint usage and posttest performance within a data-driven tutoring system. This work is novel in that it leveraged interface alone to address the help avoidance problem. However, this work did not seek to regulate students’ help-seeking, rather we sought to make unsolicited hints more effective through changes in their delivery. The Assertions hint interface made the intelligent tutor more effective, significantly improving unsolicited hint usage for all students. We further demonstrated aptitude-treatment interaction effects where students with low prior proficiency receiving Assertions performed better in the posttest, in terms of both time and solution length. Our cluster analysis shows that the students with low prior knowledge who received Assertions demonstrate more productive persistence in that they exerted more persistent effort even when failing during training, and used a higher proportion of unsolicited hints, but performed better on the posttest than their low peers who received Messages.

There are three main limitations to this study. Assertions were provided significantly more frequently than Messages. Assertions did not seem to have a negative impact on learning, but rather leveled the playing field for students with low prior proficiency. However, our analyses demonstrated that it was not hint frequency but the Assertions interface alone that improved hint usage. The second limitation was that Assertions appeared randomly, and were not adapted to individual students. Our results confirm our hypothesis that the Assertions have a differential impact for students with different incoming proficiency, suggesting that there may be benefits to using individual factors to determine when to provide Assertions. A third limitation arises from splitting students into two prior proficiency groups. While some studies investigate finer-grained partitions, e.g. low, medium/average, and high groups \cite{kardan2015providing}, we refrained from doing so to maintain sufficiently high sample sizes within each group. 

This study was a necessary first step to identify a hint interface that could solve the help avoidance problem. Future work could study the generalizability of this transformative new genre of unsolicited hints that use the design principles of contiguity, attention, and expectation to increase hint immediacy and persuasion to reduce help avoidance in other tutors. Within our tutor, we plan to apply reinforcement learning and other machine learning techniques to derive an adaptive policy to decide when and if Assertions should be provided to individual students. Since Assertions promote productive persistence among students with low prior knowledge, we also plan to develop a model that provides Assertions when the tutor detects or predicts unproductive behaviors \cite{maniktala2020dc, maniktala2020extending}.

\section{Acknowledgement}
This material is based upon work supported by the National Science Foundation under Grant No. 1726550, ``Integrated Data-driven Technologies for Individualized Instruction in STEM Learning Environments.", led by Min Chi and Tiffany Barnes. We would like to thank Nicholas Lytle (nalytle@ncsu.edu) for suggesting edits in the introduction section to enhance its clarity.

\bibliographystyle{spmpsci}
\bibliography{refs.bib}

\clearpage
\appendix
\section{: Unsolicited Hint Metrics for each prior proficiency group}
\label{appendix_hint_usage}
\FloatBarrier
\begin{table}[H]
\centering
\centering
\begin{adjustwidth}{-0.75cm}{}
\begin{tabular}{c|cccccc}
 \cline{2-7} 
Prior  & \multicolumn{2}{c}{\#Given}  & \multicolumn{2}{c}{HJR}   & \multicolumn{2}{c}{HNR}   \\ \cline{2-7} 
Profic-    & Assertions    & Messages      & Assertions  & Messages    & Assertions  & Messages    \\
iency                        & Mean (SD)     & Mean (SD)     & Mean (SD)   & Mean (SD)   & Mean (SD)   & Mean (SD)   \\ \rowcolor[HTML]{EFEFEF}\hline
Low                     & 48.92 (11.76)* & 35.71 (10.52) & 0.93 (0.09)* & 0.63 (0.18) & 0.83 (0.08)* & 0.61 (0.17) \\
High                    & 48.67 ( 7.80)* & 30.93 (14.00) & 0.92 (0.07)* & 0.63 (0.15) & 0.82 (0.10)* & 0.62 (0.16) \\ \rowcolor[HTML]{EFEFEF} \bottomrule
All                     & 48.82 (9.85)*  & 32.74 (10.64) & 0.93 (0.07)* & 0.63 (0.18) & 0.82 (0.09)* & 0.62 (0.17) \\ 
\end{tabular}
\end{adjustwidth}

\end{table}

\section{: Comparison of Posttest Accuracy between the two conditions}
\label{apx_accuracy}
\begin{table}[H]
\centering
\begin{tabular}{r|cc}
\hline
\begin{tabular}[c]{@{}l@{}}Prior\\Profi-\\ciency\end{tabular} & Assertions     & Messages        \\
& Mean (SD) & Mean (SD)\\\hline
Low                                                           & 0.74 (0.10)& 0.72 (0.08) \\\rowcolor[HTML]{EFEFEF}
High                                                          & 0.75 (0.09)& 0.73 (0.08)  \\
All                                                           & 0.74 (0.10) & 0.73 (0.08)    
\end{tabular}
\end{table}

\end{document}